\documentclass[a4paper,conference]{IEEEtran}

\usepackage{graphicx}
\usepackage{amsmath,amssymb} 

\usepackage{algorithm}

\usepackage[noend]{algpseudocode}

\usepackage{epsfig}

\usepackage{caption}

\ifCLASSINFOpdf
\else
\fi
\usepackage{amsmath}

\begin{document}

\title{Identity-aware Facial Expression Recognition in Compressed Video}

\author{\IEEEauthorblockN{Xiaofeng Liu$^{1,5\dag}$, Linghao Jin$^{3\dag}$, Xu Han$^{3}$, Jun Lu$^{1*}$, Jane You$^4$, Lingsheng Kong$^2$}\\
\IEEEauthorblockA{$^1$ Beth Israel Deaconess Medical Center and Harvard Medical School, Boston, MA, USA\\$^2$ Changchun Institute of Optics, Fine Mechanics and Physics, Chinese Academy of Sciences, CAS, Changchun, China\\$^3$ Johns Hopkins University, Baltimore, MD, USA\\$^4$ Dept. of Computing, Hong Kong Polytechnic University, Hung Hom, Hong Kong\\$^5$ Fanhan Tech. Inc., Suzhou, Jiangsu, China.\\
\dag Contribute equally. *Corresponding Author.}
}

\maketitle

\begin{abstract}
This paper targets to explore the inter-subject variations eliminated facial expression representation in the compressed video domain. Most of the previous methods process the RGB images of a sequence, while the off-the-shelf and valuable expression-related muscle movement already embedded in the compression format. In the up to two orders of magnitude compressed domain, we can explicitly infer the expression from the residual frames and possible to extract identity factors from the I frame with a pre-trained face recognition network. By enforcing the marginal independent of them, the expression feature is expected to be purer for the expression and be robust to identity shifts. We do not need the identity label or multiple expression samples from the same person for identity elimination. Moreover, when the apex frame is annotated in the dataset, the complementary constraint can be further added to regularize the feature-level game. In testing, only the compressed residual frames are required to achieve expression prediction. Our solution can achieve comparable or better performance than the recent decoded image based methods on the typical FER benchmarks with about 3$\times$ faster inference with compressed data.
\end{abstract}

\IEEEpeerreviewmaketitle

\section{Introduction}

Considering the natural dynamic property of the human face expression, many works propose to explore spatio-temporal features of facial expression recognition (FER) from the videos. Although the multi-frame sequence can inherit richer information and the temporal-correlation between the consecutive frames can usually be helpful for the facial expression recognition, the video also introduced significantly redundancy. Considering the subtle muscle movement in FER videos, the signal-to-noise ratio (SNR) can be low \cite{liu2019hard,li2020deep}.

\begin{figure}[t]
\centering
\includegraphics[width=9cm]{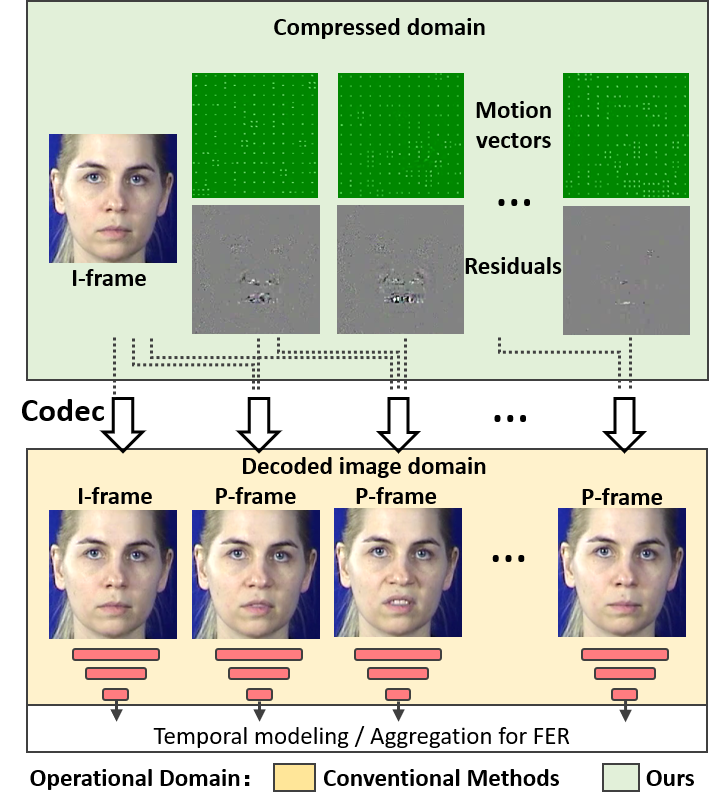}\\
\caption{Illustration of the typical video compression and the scheme of conventional FER methods, which first decode the video and then feed it into a FER network.}\label{fig:1}
\end{figure}

The conventional spatio-temporal FER networks take sequentially frames to utilize the spatial and temporal cues to represent the facial expressions \cite{li2020deep}. The recursive neural network (RNN), 3D convolutional, and non-local network can be used to derive information from sequences \cite{kim2017multi,tran2015learning,meng2019frame}. However, using deep neural networks to process many consecutive frames can be very computationally expensive when the length of video is long \cite{liu2018dependency,liu2019attention,liu2019permutation,liu2019dependency}. Actually, many methods can achieve good performance using decision-level fusion of image-based FER, which totally ignores the temporal dependency. This implies the temporal cues may hard be explored in the low SNR FER videos.

We propose that the compressed domain can be suitable for FER task for four reasons. \textbf{1)} the consecutive frames in video modality have many uninformative and repeating patterns, which may drowning the ``interesting" and ``true" signal \cite{yeo2006compressed}. \textbf{2)} the typical compression solution (e.g., MPEG-4, H.264, and HEVC) separate the video to the I frame (intracoded frames) with the first image, and follows several P frames (predictive frames) which encoded as the ``change" or ``movement" \cite{le1991mpeg}, as shown in Figure. \ref{fig:1}. The movement of face muscle is the fundamental of expression \cite{lucey2010extended}. \textbf{3)} our compressed domain exploration can also be effective since it focus on the ``true" signals rather than processing the repeatedly near-duplicates \cite{wu2018compressed,shou2019dmc}. \textbf{4)} we do not need the decode operation in real-world task, since the to-be processed data is usually transferred with the compressed format.

Moreover, the high inter-subject variations caused by identity differences in facial attributes is a long-lasting difficulty in FER community \cite{liu2017adaptive,liu2018normalized}. We note that the P frames can also incorporate the relative position of key points, which are also related to the identity \cite{tian2011facial}. The typical solution for eliminating identity from the expression representation is using the metric learning \cite{meng2017identity,liu2017adaptive} or the conditional generative adversarial network (GAN) \cite{ali2019all}. However, these works focus on the image-based FER, and most of them requires the identity label and multiple expressions of the same person, which significantly limits their applications on the in-the-wild FER dataset \cite{dhall2014emotion}.

In this work, we propose to explore the identity factor from the I frame with the pre-trained face recognizer, e.g., FaceNet \cite{schroff2015facenet}. Their embeddings are remarkably reliable, since they achieve high accuracy over millions of identities \cite{kemelmacher2016megaface}, and robust to a broad range of nuisance factors such as expression, pose, illumination and occlusion variations. Using the identity feature as the anchor, we can explicitly enforce the marginally independence of our identity and expression feature \cite{liu2019feature}.

In summary, this paper makes the following contributions.

\noindent $\bullet$ We propose to inference expression from the residual frames, which explores the off-the-shelf yet valuable expression-related muscle movement in the up to two orders of magnitude compressed domain.

\noindent $\bullet$ Targeting for the identity-aware video-based FER, the independence of expression and identity representations from P frames and I frame are enforced.

\noindent $\bullet$ The proposed IFERCV framework achieve state-of-the-art results on several video-based FER benchmarks with much faster inference. The promising performance evidenced its generality and scalability.

\section{Related Works}

\noindent{\bf{Video-based FER}}. With the fast development of deep learning \cite{che2019deep,liu2020wasserstein,liu2020unimodal,liu2020importance,liu2020severity,he2020image,han2020wasserstein,liu2020auto3d}, both the frame aggregation and spatiotemporal FER networks are developed and outperforms the conventional methods \cite{liu2017line,liu2018joint}. The frame aggregation methods can utilize the image-based FER networks by making the decision-level \cite{kahou2016emonets} or the feature-level frame-wise aggregation \cite{xu2016video}. The spatio-temporal FER networks take sequentially images to utilize both the spatial/textural and temporal information \cite{al2018deep}. However, the previous works only consider the image domain, and the spatiotemporal FER does not significantly outperforms aggregation methods \cite{li2020deep}. To the best of our knowledge, this is the first effort to investigate the compressed video FER, which is orthogonal to these advantages and can be easily added to each other \cite{he2020classification,liu2019unimodal,liu2019conservative,liu2018ordinal,liu2018data}.

\noindent{\bf{Video compression}} Usually, the video codecs separates a video into several Group Of Pictures (GOP). A GOP is composed of an I frame, and followed by several P-frames or B-frames. I-frame indicates a self-contained RGB frame with full visual representation, while the P-frame or B-frame is the inter frames which hold motion vectors and residuals w.r.t. the previous frame \cite{le1991mpeg}. \cite{zhang2016real} propose to replace the optical flow by the motion vector. However, decoding the RGB images is still necessary in this method. More recently, in the action recognition task, \cite{wu2018compressed} propose to fuse all of the available modalities in the compressed video, e.g., I-frame, residuals and motion vectors, to avoid the decoding of images. Although the expression shares some similarities with action recognition, the movement range and the utilization of I frame are essentially different. The I frame is directly used to predict the action and combine with the result of P frames \cite{wu2018compressed}, while the I frame in FER usually be the neutral face (different from the video label). Besides, the low resolution motion vector can not well encode the expression.

\begin{figure}[t]
\centering
~\\~\\\includegraphics[width=8.8cm]{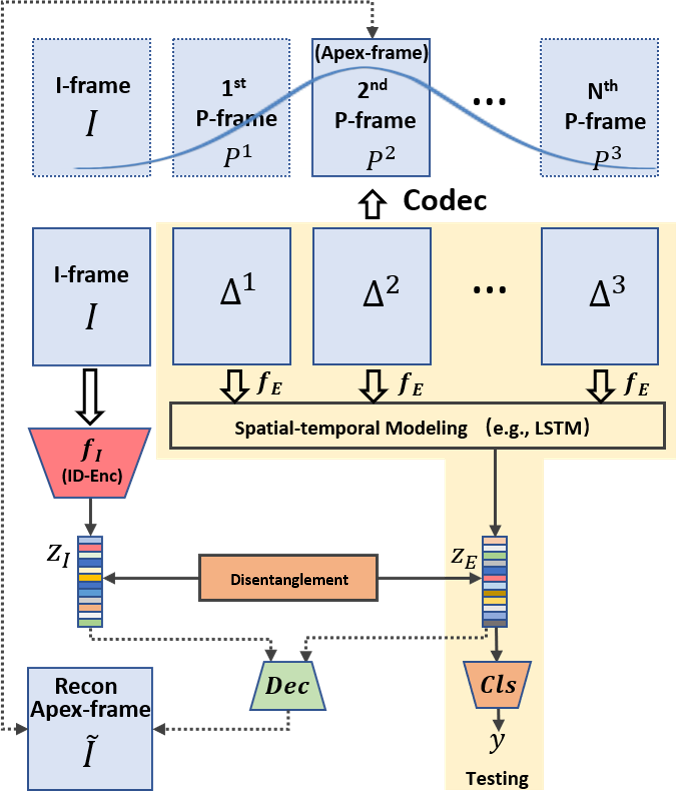}\\
\caption{The illustration of IFERCV framework for identity-aware facial expression recognition in compressed video domain.}\label{fig:2}
\end{figure}

\section{Methodology}

Our goal is to design a fast and effective video-based FER framework operates directly on the compressed domain. The overall frame work is shown in Figure. 2, which are consist of four core modules. The pre-trained identity branch and FER branch (frame embedding network $f_E$, aggregation module and Classifier) work on the I frame and undecoded P frames respectively. Then, a disentanglement module is applied to measure the dependence of identity and expression representations. Moreover, the complementary constraint can be applied when the apex frame is annotated to stabilize the early stage training.

\subsection{Modeling Compressed Representations}

To illustrate the format of input video, we choose the MPEG-4 as an example \cite{richardson2003h264}. The compressed domain has two typical frames, i.e., I frame and P frames. Specifically, the I frame $I\in\mathbb{R}^{h\times w\times 3}$ is a complete RGB image. We use $h$ and $w$ to denote its height and width respectively. Besides, the P frame at time $t$ $P^t\in\mathbb{R}^{h\times w\times 3}$ can be reconstructed with the stored offsets, called residual errors $\Delta ^t\in\mathbb{R}^{h\times w\times 3}$ and motion vectors $\mathcal{T}^t\in\mathbb{R}^{h\times w\times 2}$.

Noticing that the motion vectors $\mathcal{T}^t$ has much lower resolution, since its values within the same macroblock are identical. Considering the micro movements of facial expression in each frame, the coarse $\mathcal{T}^t$ usually not helpful for the FER. 
For P frame reconstruction $P_i^t=P_{i-\mathcal{T}_i^t}^{t-1}+\Delta_i^t$, where index all the pixels and $P^0=I$. Then, $\mathcal{T}^t$ and $\Delta ^t$ are processed by discrete cosine transform and entropy-encoded.

The typical compression algorithms are only developed to compress the file size, and the encoded format can be very different with the RGB images w.r.t. the statistical and structural properties. Therefore, a tailored processing network is necessary to accommodate the compressed format. Considering the structure of residual images $\Delta ^t$ are much simpler than the decoded images, it is possible to utilize simpler and faster CNNs $f_E:\mathbb{R}^{h\times w\times 3}\rightarrow \mathbb{R}^{512}$ to extract the feature of each frame \cite{kim2017multi,baddar2019mode}. Practically, we follow the 
CNN in the typical CNN-LSTM FER structure \cite{baddar2019mode,kim2017multi}, but with fewer layers to explore the information in $\Delta^t$. Noticing that $f_E$ is shared for all frames, and only needs to store one $f_E$ in processing.

Besides, most existing action recognition methods with compressed video \cite{wu2018compressed,shou2019dmc} independently concatenate the paired $\Delta ^t$ and $\mathcal{T}^t$ at each time step and predict an action score of each P-frame. The temporal cues and its development patterns are important for FER task \cite{li2020deep}. We simply choose the LSTM in \cite{baddar2019mode} to model the sequential development of residual frames and summarize the information to a expression feature $z_E$. Since our LSTM is applied to 512-dim features, the computation burden is largely smaller than work on the raw images. Noticing that more advanced RNN, 3D CNN or attention networks can potentially be utilized to replace our LSTM model to further boost the performance \cite{barros2016developing,zhao2018learning,kumawat2019lbvcnn}.

For the I frame with raw image format, we simply use the FaceNet \cite{schroff2015facenet} pre-trained on millions of identities \cite{kemelmacher2016megaface} as our identity feature extractor $f_{I}:\mathbb{R}^{h\times w\times 3}\rightarrow z_{I},$ where $z_{I}\in\mathbb{R}^{1024}$ denotes the identity feature. Actually, the FER videos in many datasets start from the neutral expression which can further facilitate identity recognition.

\subsection{Disentanglement and Complementary Constraint}
Configuring a dual-branch encoder to extract the separated information has been a standard protocol for disentanglement \cite{liu2019feature}. Several adversarial disentanglement works demonstrate that simply separate the input may result in the extracted feature has no meaningful information \cite{mathieu2016disentangling,liu2019feature}. The reconstruction of input can explicitly enforce the disentangled factors to be complementary to each other. However, reconstructing the video can be hugely underconstrained.

Many FER datasets follow a well-defined collection protocol, which usually starts from the neutral face and then develops to an expression. Specifically, the video in CK+ \cite{kanade2000comprehensive,lucey2010extended} consists of a sequence which shift from the neutral expression to a apex facial expression. The last frame usually be the apex frame, which has the most strong expression intensity. Actually, the image-based FER methods select the last three frames to construct their training and testing datasets. Similarly, in MMI \cite{pantic2005web}, the video frames are usually start from the neutral face and develop to the apex around the middle of video, and returning back to the neutral in the end of video. Noticing that the apex frame (i.e., last frame in CK+ or middle frame in MMI) can clearly incorporate both the identity and expression information. Therefore, we are possible to utilize the apex frame as a reference of reconstruction, and simply apply the $\mathcal{L}_1$ loss.  

\begin{equation} \label{eq:l1}
\mathcal{L}_1 = ||I_{Apex}-\hat{I}_{Apex}||_2^2
\end{equation}where $\hat{I}_{Apex}=Dec(z_{I},z_{E})$. We note that the complementary constraint is not a necessity in our framework, since the FER loss takes a large weight in the FER branch. It requires to maintain sufficient information w.r.t. expression and not easy to have nothing meaningful. But the complementary constraint does helpful for the convergence in the early stage. When the apex is annotated, we only need to decode the apex frame in the decoded image domain in the start of few training epochs.

\subsection{Overall objectives}

 We have three to be minimized objectives, i.e., cross-entropy loss, disentanglement loss \cite{liu2019feature} and $\mathcal{L}_1$ loss, which works collaboratively to update each module. The expression classification is the main task of the FER. We choose the typical cross-entropy loss $\mathcal{L}_{CE} = -\sum_{c=1}^C y_{c}\log(Cls(z_E)_{c})$ to ensure $z_E$ contains sufficient expression information and finally have a good performance on $C$-class expression classification. We use $y_{c}$ and $Cls(z_E)_{c}$indicate the $c^{th}$ class probability of the label and classifier softmax predictions respectively. Since the FER branch can be updated with all of the losses, we assign the balance parameter $\alpha\in[0,1]$ and $\beta\in[0,1]$ to disentanglement loss \cite{liu2019feature} and $\mathcal{L}_1$ loss minimization objectives respectively. We note that only the FER branch, i.e., $f_E$, LSTM and Cls, is used for testing.

\section{Experiments}\label{sect:exp}

In this section, we first detail our experimental setup, present quantitative analysis of our model, and finally compare it with state-of-the-art methods.

\begin{table}[t]
~\\~\\\caption{Comparison of various methods on the CK+ dataset in terms of average recognition accuracy of seven expressions. Note that in order to make the comparison fair, we do not consider image-based and 3D geometry based experiment setting and models \cite{liu2017adaptive,meng2017identity,liu2019hard}.}\vspace{+10pt}
	\label{tab:ck_all}
	\resizebox{\columnwidth}{!}{
	\begin{tabular}{| l | c | c | c |}
		\hline
		\textbf{Method} & \textbf{Accuracy}  & \textbf{Landmarks} & \textbf{Test speed} \\
		\hline

		STM-ExpLet (2014) \cite{liu2014learning} 	& 94.19 & $\times$       &-\\
		LOMo  (2016) \cite{sikka2016lomo}			& 95.10 & $\checkmark$	 & -\\
		DTAGN  (2015) \cite{jung2015joint} 			& 97.25 & $\checkmark$   & - \\
		PHRNN-MSCNN (2017) \cite{zhang2017facial}   & 98.50 & $\checkmark$   & -\\
		(N+M)-tuplet (2018) \cite{liu2018adaptive}   & 93.90 & $\checkmark$   & 12fps  \\
		C3D-GRU (2019) \cite{lee2019visual} & 97.25 &$\times$		 & - \\
		CTSLSTM (2019) \cite{hu2019video}	& 93.9 & $\checkmark$       & - \\ 	
		SC (2019) \cite{verma2019facial}		& 97.60 & $\checkmark$      & - \\ 	
		G2-VER (2019) \cite{albrici2019g2}			& 97.40 & $\times$       & - \\ 			
		LBVCNN (2019) \cite{kumawat2019lbvcnn}		& 97.38 & $\times$       & - \\ \hline		\hline 

Mode variational LSTM (2019) \cite{baddar2019mode} & 97.42 & $\times$       & 11fps   \\\hline

IFERCV & 97.44 & $\times$       & \textbf{35fps}    \\
IFERCV+Adversarial disentanglement\cite{liu2019feature}   & 98.38 & $\times$       & \textbf{35fps}  \\
IFERCV-$\hat{I}$  & 97.16 & $\times$       & \textbf{35fps}   \\
IFERCV+$\mathcal{T}^t$  & 97.85 & $\times$       & 29fps     \\\hline

	\end{tabular}	}

\end{table}

\subsection{Description of the datasets}

\noindent\textbf{CK+ Dataset} is a widely accepted FER benchmark \cite{li2020deep,kumawat2019lbvcnn}. The video is collected in a restricted environment, in which the participate subjects are facing to the recorder with the empty background. The video in CK+ consists of a sequence which shift from the neutral expression to a apex facial expression. The last frame usually be the apex frame, which has the most strong expression intensity. Following the previous works, we use subject independent 10-folds cross validation \cite{liu2014learning,li2020deep,liu2018adaptive}.

Many FER datasets follow a well-defined collection protocol, which usually starts from the neutral face and then develops to an expression. Specifically,  Actually, the image-based FER methods select the last three frames to construct their training and testing datasets. Similarly, in MMI \cite{pantic2005web}, the video frames are usually start from the neutral face and develop to the apex around the middle of video, and returning back to the neutral in the end of video. Noticing that the apex frame (i.e., last frame in CK+ or middle frame in MMI) can clearly incorporate both the identity and expression information. Therefore, we are possible to utilize the apex frame as a reference of reconstruction, and simply apply the $\mathcal{L}_1$ loss.

\begin{figure}[t]
\centering
~\\~\\\includegraphics[width=7cm]{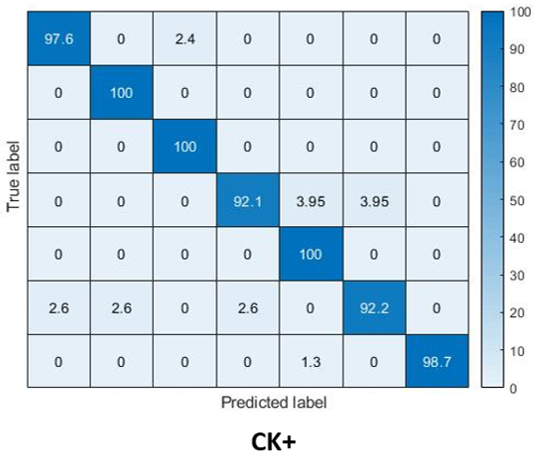}\\
\caption{Confusion matrix of IFERCV on CK+ datasets.}\label{fig:4} 
\end{figure}

\noindent\textbf{AFEW Dataset} is more close to the uncontrolled real-world environment. It is consist of the video clips of movies \cite{perveen2018spontaneous}. The video in AFEW has the spontaneous facial expression. The AFEW has seven expressions: anger, disgust, fear, happiness, sadness, surprise and neutral. Following the evaluation protocol in EmotiW \cite{dhall2017individual}, there are training, validation and testing sets. Since its testing label is not available, we follow the previous work to use the validation set for comparison \cite{baddar2019mode}. Noticing that the validation set is not used in training stage for parameter or hyper-parameter tuning.

\subsection{Evaluation and ablation study}

\noindent\textbf{Results on CK+ dataset.} The 10-fold cross-validation performance of our proposed method is shown in the Table 1. For fair comparison, the image-based experiment settings are not incorporated in the tables. Besides, only the state-of-the-art (SOTA) accuracy obtained by the single-models (non-ensemble model) are listed.

Many models, e.g., DTAGN \cite{jung2015joint}, LOMo \cite{sikka2016lomo}, PHRNN-MSCNN \cite{zhang2017facial}, CTSLSTM \cite{hu2019video} and SC \cite{verma2019facial}, achieved the SOTA performance by utilizing the facial landmarks. However, this operation highly relies on the fine-grained landmark detection, which itself is a challenging task \cite{li2020deep,liu2019hard}, and unavoidably introduced additional computation.  

Based on the mode variational LSTM \cite{baddar2019mode}, our IFERCV achieves the SOTA result without the facial landmarks, 3D face models or optical flow. It worth noticing that the much simpler CNN encoding network makes more residual frames can be processed parallel than \cite{baddar2019mode}. Moreover, our efficiency is also benefit from avoiding the decompress of the video. Since the videos are typically stored and transmitted with the compressed version, and the residuals are off-the-shelf. As a result, the proposed compressed domain IFERCV can speed up the testing about 3 times over \cite{baddar2019mode}, and achieve better accuracy.

Besides, \cite{liu2018adaptive} is a typical metric-learning based identity removing method. Our solution can significantly outperform it with respect to both speed and accuracy. Actually, the sampling of tuplets usually makes the training not scalable \cite{liu2019hard}, while our identity eliminating scheme is concise and effective.

When we remove some modules from our framework, the performances have different degrees of decline. We use -$\hat{I}$ to denote the IFERCV without complementary constraint. The result also implies that the identity can be well encoded by the face recognition network. 

We can also follow the action recognition method \cite{shou2019dmc} to concatenate the motion vector and residual as the input, and denote as IFERCV+$\mathcal{T}^t$. However, we do not achieve significant performance on all datasets, but the inference speed in testing can be slower. This maybe related to the coarse resolution of the motion vector can not well describe the fine-grain muscle movement of the face.

The confusion matrix of our proposed IFERCV method on the CK+ is reported in Figure 4 (left). The accuracy for the expression class  happiness, disgust, angry, surprise and contempt are almost perfect.

\begin{table}[t]
~\\~\\\caption{Comparison of various methods on the AFEW dataset in terms of average recognition accuracy of seven expressions. *optical flow is used.}\vspace{+10pt}
	\label{tab:afew}
	\resizebox{\columnwidth}{!}{
	\begin{tabular}{| l | c | c | c |}
		\hline
		\textbf{Method} & \textbf{Model type}  & \textbf{Accuracy} & \textbf{Test speed} \\
		\hline

CNN-RNN (2016) \cite{fan2016video}& 45.43 &Dynamic&-\\
Undirectional LSTM (2017) \cite{vielzeuf2017temporal} & 48.60&Dynamic&-\\
HoloNet (2016) \cite{yao2016holonet} & 44.57&Static&-\\
DSN-HoloNet (2017) \cite{hu2017learning}  &46.47&Static&-\\
DenseNet-161 (2018) \cite{liu2018multi} & 51.44&Static&-\\
DSN-VGGFace (2018) \cite{fan2018video}  &48.04&Static&-\\
FAN (2019) \cite{meng2019frame} &51.18&Static&-\\
CTSLSTM (2019) \cite{hu2019video} & 51.2 &Dynamic&-\\
C3D-GRU (2019) \cite{lee2019visual} & 49.87&Dynamic&-\\
DSTA (2019)* \cite{pan2019deep} & 42.98&Dynamic&-\\
E-ConvLSTM (2019)* \cite{miyoshi2019facial} & 45.29&Dynamic& 4fps   \\\hline\hline
Mode variational LSTM (2019) \cite{baddar2019mode} & 51.44& Dynamic&  11fps  \\\hline

IFERCV  & 51.62& Dynamic&   \textbf{34fps}  \\
IFERCV+Adv Disentanglement\cite{liu2019feature}  & 52.01& Dynamic&  \textbf{34fps}   \\
IFERCV+$\mathcal{T}^t$  & \textbf{51.86} & Dynamic&   30fps   \\\hline

	\end{tabular}}
\end{table}

\noindent\textbf{Results on AFEW dataset.} The evaluation of the proposed IFERCV on AFEW dataset is shown in Table 4. We note that only the SOTA accuracy obtained by the single-models (non-ensemble model) are listed for fair comparison. Besides, the audio modality in AFEW can be used to boost the recognition performance \cite{fan2017dynamic,fan2016video,vielzeuf2017temporal}. We note that we only focus on the image compression in this paper, and the audio/video data are stored in separate tracks, but the additional modality can also potentially to be add on our framework following the multi-modal methods\cite{fan2017dynamic,fan2016video}.

With the simplified Mode variational LSTM-based \cite{baddar2019mode} backbone, the exploration in the compressed domain can achieve comparable or even better recognition performance. More promisingly, our IFERCV can also achieve real-time processing for the uncontrolled environment, which evidenced its generality. We note that the typical time resolution in FER is 24fps \cite{kim2017multi}.

\begin{figure}[t]
\centering
~\\~\\\includegraphics[width=7cm]{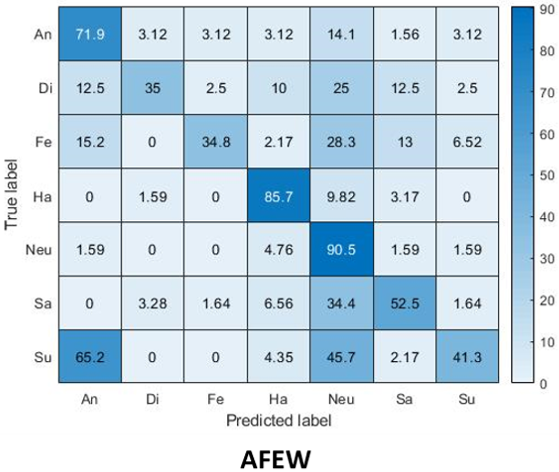}\\
\caption{Confusion matrix of IFERCV on AFEW datasets.}\label{fig:5} 
\end{figure}

Some of the works propose to improve the image-based FER networks and combine the frame-wise scores for video-based FER \cite{yao2016holonet,hu2017learning,liu2018multi,fan2018video,meng2019frame}. \cite{yao2016holonet,hu2017learning} input both the LBP maps and the image to the CNNs. \cite{hu2017learning} and \cite{fan2018video} utilize the additional supervision on intermediate layers. The image-based FER methods \cite{liu2018multi} achieves high performance, but \cite{liu2018multi} uses a very deep network DenseNet-161 and pretrains it on the private Situ dataset. Moreover, \cite{liu2018multi} utilize the sophisticated post-processing. Actually, an intuition of statistic based solution is to avoid LSTM and speed up the processing. However, with the super deep and complicated structure, their processing can be much slower than our solution.  

Both \cite{fan2016video} and \cite{vielzeuf2017temporal} use VGGFace as the backbone of $f_E$ and a RNN model with LSTM units. They target to capture the temporal dynamic cues of the videos. Moreover, \cite{hu2019video,lee2019visual,pan2019deep} also propose to modify the LSTM model for the spatial-temporal modeling. However, all of the above solutions are applied to the decoded space, which requires decoding processing and needs to handle much more complicated data. Overall, the proposed IFERCV can improve the testing speed by a large margin and can achieves the SOTA accuracy as the previous models.

\section{Conclusion}

In this work, we propose to explore the facial expression cues directly on the compressed video domain. We are motivated by our practical observation that the facial muscle movements can be well encoded in the residual frames which can be informative and free of cost. Besides, the video compression can reduce the repeating boring patterns in the videos, which rendering the representation to be robust. The increased relevance and reduced complexity or redundancy in FER videos make computation much more effective. We extract the identity and expression factor from the I frame and P frame respectively, and explicitly enforce their independence with disentanglement regularization \cite{liu2019feature}. When the apex frame label is available in training, the complementary constraint can further stabilize the training. In three video-based FER benchmarks, our IFERCV can improve the performance without the additional identity, face model or facial landmarks labels. The processing speed of the test stage is promising for real-time FER.

\section{Acknowledgements}

This work was supported by the Jangsu Youth Programme [BK20200238], National Natural Science Foundation of China, Younth Programme [grant number 61705221], NIH [NS061841, NS095986], Fanhan Technology, and Hong Kong Government General Research Fund GRF (Ref. No.152202/14E) are greatly appreciated.

\bibliographystyle{IEEEtran}
\bibliography{IEEEabrv.bib}

\end{document}